# U-Net Kalman Filter (UNetKF): An Example of Machine Learning-assisted Ensemble Data Assimilation

**Feiyu Lu[1, 2, *]**

[1]University Cooperation of Atmospheric Research

[2]National Oceanic and Atmospheric Administration/Geophysical Fluid Dynamics Laboratory

[*]Previously Princeton University

Corresponding author: Feiyu Lu (Feiyu.lu@noaa.gov)

**Key Points:**

- Machine learning methods are used to improve traditional ensemble-based data assimilation methods.
- U-Net is trained to predict the localized ensemble error covariances.
- UNetKF outperforms EnKF with small to moderate ensemble sizes in a 2-layer quasi-geostrophic model.

## Abstract

Machine learning techniques have seen a tremendous rise in popularity in weather and climate sciences. Data assimilation (DA), which combines observations and numerical models, has great potential to incorporate machine learning and artificial intelligence (ML/AI) techniques. In this paper, we use U-Net, a type of convolutional neutral network (CNN), to predict the localized ensemble covariances for the Ensemble Kalman Filter (EnKF) algorithm.

Using a 2-layer quasi-geostrophic model, U-Nets are trained using data from EnKF DA experiments. The trained U-Nets are then used to predict the flow-dependent localized error covariance matrices in U-Net Kalman Filter (UNetKF) experiments, which are compared to traditional 3-dimensional variational (3DVar), ensemble 3DVar (En3DVar) and EnKF methods. The performance of UNetKF can match or exceed that of 3DVar, En3DVar or EnKF. We also demonstrate that trained U-Nets can be transferred to a higher-resolution model for UNetKF implementation, which again performs competitively to 3DVar and EnKF, particularly for small ensemble sizes.

## Plain Language Summary

Data assimilation has been widely used in weather and climate applications to combine information from observations and numerical model simulations. One common type of modern data assimilation method is the ensemble Kalman filter and its variants, which have been used in both research and operations in the fields of weather and climate. This study will incorporate deep learning methods to complement and improve the ensemble Kalman filter algorithm. More specifically, we will train a convolutional neural network to predict the error statistics of a model ensemble, which will reduce the computational costs of the ensemble Kalman filter. This approach is tested with a quasi-geostrophic dynamic model to demonstrate its feasibility in more advanced weather and climate applications.

## 1. Introduction

Data assimilation (DA) has been used extensively in the fields of geosciences, including meteorology and climate science, to estimate present and historical states of the earth system components and provide initial conditions for numerical weather and climate predictions. Since its introduction (Evensen, 1994), Ensemble Kalman Filter (EnKF) and its variants have been widely used in weather and climate applications (Houtekamer & Mitchell, 1998; Anderson, 2001; Hunt et al., 2007; S. Zhang et al., 2007; Houtekamer & Zhang, 2016). Ensemble-based Kalman filters improve upon the previous Optimal Interpolation (OI) or 3-dimentional variational (3DVar) DA methods by using an ensemble of model simulations to estimate the real-time flow-dependent model uncertainty and error statistics. The wide adoption of ensemble-based Kalman filters is facilitated by its minimal changes to existing model code and its flexibility in implementation for parallel computation (Anderson, 2003; S. Zhang et al., 2007; Anderson et al., 2009). One persistent source of error for ensemble-based Kalman filters is the sampling errors due to limited ensemble sizes, especially for computationally expensive dynamic models. Numerous schemes to mitigate this issue have been proposed and refined over the decades, such as localization and inflation methods (F. Zhang et al., 2004; Buehner & Charron, 2007; Anderson, 2007, 2009). Although it has been shown that modest ensemble sizes on the order of tens can lead to successful state estimation in global climate models (Lu et al., 2020; S. Zhang et al., 2007), this computational constraint will only worsen when model complexity and resolutions further increase.

In recent years, the development of artificial intelligence (AI) and machine learning (ML) methods has pointed to potentially exciting advancement for weather and climate research, particularly for DA. The potential synergy between DA and ML has sparked a wide variety of studies that try to introduce AI/ML into DA, or vice versa (Boukabara et al., 2021). For ensemble-based data assimilation specifically, most studies attempted to mitigate the ensemble size limitation by using ML methods to either construct or expand the model ensemble. For examples, Grooms (2021) used variational autoencoders (VAEs) to generate model analogs for the ensemble optimal interpolation (EnOI) method. Penny et al. (2022) proposed the use of recurrent neural networks (RNNs) as surrogate models to estimate forecast error statistics and achieve pure data-driven state estimation. The rapid advancement of ML-based weather and climate models (Bi et al., 2023; Kochkov et al., 2023; Lam et al., 2023) points to an exciting future where very large ensembles can be efficiently generated for ensemble data assimilation using these data-driven models. Other examples of DA-related ML applications include Bonavita & Laloyaux (2020) and Farchi et al. (2021) , which explored the use of machine learning methods to estimate and correct the model error in 4-dimensional variational (4DVar) DA systems, and Hatfield et al. (2021) that assessed the ability of neural network (NN) emulators to provide tangent-linear and adjoint models for 4DVar. Chen et al. (2022) applied a similar model bias correction technique with NOAA's forecast model and showed that it could be applied iteratively to improve forecast skill.

In this paper, we will test the feasibility of using U-Net, a class of convolutional neural networks (CNN), to predict ensemble error covariances for EnKF. We will call this approach for ML-assisted ensemble DA the Ensemble U-Net Kalman Filter (UNetKF). The principles in this paper should be applicable to multiple earth system model components.

## 2. Model and experiment design

The numerical mode code used in this study is $pyqg$, a Python library that models quasi-geostrophic (QG) systems using pseudo-spectral methods. It has been used to showcase an open-source framework for training, testing, and evaluating data-driven subgrid parameterization in an ocean model (Ross et al., 2023). QG models provide a reasonable simplification to general circulation models (GCM). In this study, we use an idealized two-layer model from $pyqg$. Model parameters include the layer thickness ratio (upper/lower) $delta = 0.05$, upper layer zonal velocity $U_1 = 0.025\ m/s$, lower layer zonal velocity $U_2 = 0.0\ m/s$, the gradient of the planetary vorticity (Coriolis parameter) $\beta = 5 \times 10^{-12}\ (ms)^{-1}$, and bottom drag coefficient $r_{ek} = 3.5 \times 10^{-8}\ s^{-1}$. Given the small thickness ratio, the spatial and temporal variability are different between the two layers, akin to an upper ocean mixing layer and a subsurface ocean intermediate-deep layer.

We configure the two-layer QG model with a double-periodic square domain of 1000 km. An idealized biased-model framework is used, where a 50-year “nature run” or “truth” simulation is performed with a $128 \times 128$ grid, or a resolution of $\sim 7.8\ km$. “Observations” of potential vorticity from both layers are sampled every 10 days from the “truth” at 50 random locations. The DA frequency is also set at 10 days for all experiments. The observation network is designed to loosely emulate real-world observation networks, in which the locations of observations normally change with time. The standard deviations of the random "observation" errors are $10^{-5}$ and $5 \times 10^{-7}\ m^2\ s^{-1}\ K\ kg^{-1}$ for the upper and lower layers, respectively. These errors are arbitrarily chosen to match primarily their respective representative errors, namely the typical spatial standard deviations within a small domain in the high-resolution "truth" that might only be represented by a single grid-point in the low-resolution prediction model. Such representative errors arise when we assimilate highly localized in situ observations into gridded model locations where each gridpoint represent certain area averages. Other choices of data assimilation frequency, observation frequency, density and error are also tested in additional experiments, and the conclusions are not substantially changed. Two other versions of the two-layer QG model, including a higher-resolution QG-H(igh) with a $64 \times 64$ grid and a lower-resolution QG-L(ow) with a $32 \times 32$ grid, are used to represent different classes of "GCMs" for the "real world".

The DA experiments are evaluated based on the root mean squared errors (RMSE) of potential vorticity between the daily model analyses and the "truth", which is linearly interpolated to match the resolutions of the DA experiments. The significance of the differences in RMSE between different DA experiments are tested using the standard t-test and a 95% confidence level, where the degree of freedom (DOF) of each time series is adjusted based on its autocorrelation timescale (Panofsky & Brier, 1968):

$$DOF = \frac{N\Delta t}{2T_e} \qquad (1)$$

$N$ is the length of the time series. $\Delta t$ is the timestep interval. $T_e$ is the e-folding decay time of autocorrelation, namely the time when the autocorrelation coefficient of the time series drops to $1/e$.

## 3. Data assimilation methods

3Dvar has been a popular DA method for decades (Derber & Rosati, 1989; Lorenc, 1986), and is still used by operational ocean reanalysis systems (Waters et al., 2015; Zuo et al., 2017). The solution (analysis $\boldsymbol{x}^a$) of 3Dvar is obtained by minimizing the cost function (Lorenc, 1986)

$$\mathcal{J}(\boldsymbol{x}) = \frac{1}{2}(\boldsymbol{x} - \boldsymbol{x}^b)^T \boldsymbol{B}^{-1}(\boldsymbol{x} - \boldsymbol{x}^b) + \frac{1}{2}(\boldsymbol{y}^o - \boldsymbol{H}(\boldsymbol{x}))^T \boldsymbol{R}^{-1}\big(\boldsymbol{y}^o - \boldsymbol{H}(\boldsymbol{x})\big) \quad (2)$$

where $\boldsymbol{y}^o$ is the observation, $\boldsymbol{x}^b$ model forecast, $\boldsymbol{H}$ the observational operator from model to observational space, $\boldsymbol{R}$ the observational error covariance matrix, and $\boldsymbol{B}$ the climatological background or prior covariance matrix. To get $\nabla_x \boldsymbol{\mathcal{J}}(\boldsymbol{x^a}) = 0$, we can write in incremental form $\boldsymbol{x^a} = \boldsymbol{x^b} + \boldsymbol{\delta x}$, linearize $\boldsymbol{H}(\boldsymbol{x^a}) = \boldsymbol{H}(\boldsymbol{x^b} + \boldsymbol{\delta x}) \approx \boldsymbol{H}(\boldsymbol{x^b}) + \boldsymbol{H}(\boldsymbol{\delta x})$, and get

$$\delta \boldsymbol{x} = (\boldsymbol{B}^{-1} + \boldsymbol{H^T}\boldsymbol{R}^{-1}\mathrm{H})^{-1}\boldsymbol{H}^T\boldsymbol{R}^{-1}(\boldsymbol{y}^o - \boldsymbol{H}(\boldsymbol{x}^b)) \qquad (3)$$

or the equivalent optimal interpolation solution

$$\delta \boldsymbol{x} = \boldsymbol{B}\boldsymbol{H}^T(\boldsymbol{R} + \boldsymbol{H}\boldsymbol{B}\boldsymbol{H^T})^{-1}(\boldsymbol{y}^o - \boldsymbol{H}(\boldsymbol{x}^b)) \qquad (4)$$

The background or prior covariance $\boldsymbol{B}$ for 3Dvar is calculated based on separate 500-year control simulations following Parrish & Derber (1992). After testing various forecast length, the consecutive 60-day forecast error covariances over the 500 years are chosen as the 3Dvar background covariance $\boldsymbol{B}$. The shape of the covariance structure does not change significantly over different forecast windows, while the magnitude does increase with forecast window length. Similar 3Dvar results can also be achieved by empirically tuning the background covariances from shorter forecast windows.

The ensemble 3Dvar (En3Dvar) takes the ensemble mean of multiple independent 3Dvar analyses to achieve better performance.

The EnKF is an approximation of the Kalman filter. It uses an ensemble of model realizations to propagate and estimate the model forecast error statistics. In its traditional form with perturbed observations, the EnKF can also be written in incremental form as

$$\delta \boldsymbol{x}_i = \mathbf{K}(\boldsymbol{y}^o - \boldsymbol{y_i}^b), \quad i = 1, \ldots, N, \text{ where } \boldsymbol{y_i}^b = \boldsymbol{H}(\boldsymbol{x}_i^b) + \boldsymbol{\varepsilon}_i \qquad (5)$$

$$\mathbf{K} = \boldsymbol{B}_{ens}\boldsymbol{H}^T(\boldsymbol{R} + \boldsymbol{H}\boldsymbol{B}_{ens}\boldsymbol{H^T})^{-1} \qquad (6)$$

where $\delta \boldsymbol{x}_i$ are the DA increments, $\boldsymbol{K}$ the Kalman gain, $\boldsymbol{y^o}$ the observation, $\boldsymbol{x_i}^b$ the model forecast, $\boldsymbol{\varepsilon}_i$ the added perturbations on the model forecasts, $\boldsymbol{y_i}^b$ the perturbed model forecasts in the observation space, and $N$ is the ensemble size, and $i$ the individual ensemble member. $\boldsymbol{H}$ and $\boldsymbol{R}$ are the same matrices described above in 3Dvar, and $\boldsymbol{B}_{ens}$ **is** the model error covariance calculated from the ensemble forecast perturbations relative to the ensemble forecast mean:

$$\boldsymbol{B}_{ens} = \frac{1}{N-1}\sum_1^N(\boldsymbol{x_i}^b - \overline{\boldsymbol{x_i}^b})(\boldsymbol{x_i}^b - \overline{\boldsymbol{x_i}^b})^T, \quad i = 1, \ldots, N. \qquad (7)$$

For the UNetKF implementation, we take the same EnKF equation (4) and (5), but replace the ensemble model error covariance $\boldsymbol{B}_{ens}$ with the U-Net predicted covariance $\boldsymbol{B}_{UNet}$ in the Kalman gain.

$$\mathrm{K} = \boldsymbol{B}_{UNet}\boldsymbol{H}^{T}(\boldsymbol{R} + \boldsymbol{H}\boldsymbol{B}_{UNet}\boldsymbol{H}^{\boldsymbol{T}})^{-1} \qquad (8)$$

where $\boldsymbol{B}_{UNet}$ is predicted by U-Net from either the ensemble-mean model forecast $\overline{\boldsymbol{x}_{\iota}}^{b} = \frac{1}{N}\sum_{1}^{N}\boldsymbol{x}_{i}{}^{b}$ when multiple ensemble members are used, or the model forecast $\boldsymbol{x}^{b}$ when UNetKF uses a single deterministic member. The details of the U-Net, including its architecture and training, will be described in the next section.

Covariance localization is applied to all DA methods by using the Schur product of the covariance matrix ($\boldsymbol{B}/\boldsymbol{B}_{ens}/\boldsymbol{B}_{UNet}$) and a localization matrix $\boldsymbol{W}$ of the same size, in which the values are calculated based on the Gaspari & Cohn (1999) function and the physical distance between each pair of model gridpoints.

## 4. U-Net

### 4.1. Architecture

A convolutional neural network (CNN) is a class of artificial neural networks that are commonly used to analyze images. The ability to identify and extract features from images makes the CNN a great fit for certain geophysical applications. The CNN and its variants have been used in subgrid parameterization and machine learning weather forecasting (Rasp & Thuerey, 2021; Sønderby et al., 2020; Weyn et al., 2020). The U-Net is a type of CNN that was first used for biomedical image segmentation (Ronneberger et al., 2015), and later adopted by remote sensing and geographic information system research (John & Zhang, 2022; Lagerquist et al., 2021). U-Net can not only extract features from an image, but also project such features back to the size of the original image for segmentation or classification purposes. Another key advancement of U-Net is the skip connections between higher-dimensional layers (grey arrows in Figure 1) that help the network learn higher-resolution features from the input data before further dimension reduction. A schematic representation of the U-Net architecture can be found in Ronneberger et al. (2015) and is reproduced with updated parameters for this study in Figure 1. The U-Net models can be built using existing deep learning libraries, and we used PyTorch for this study.

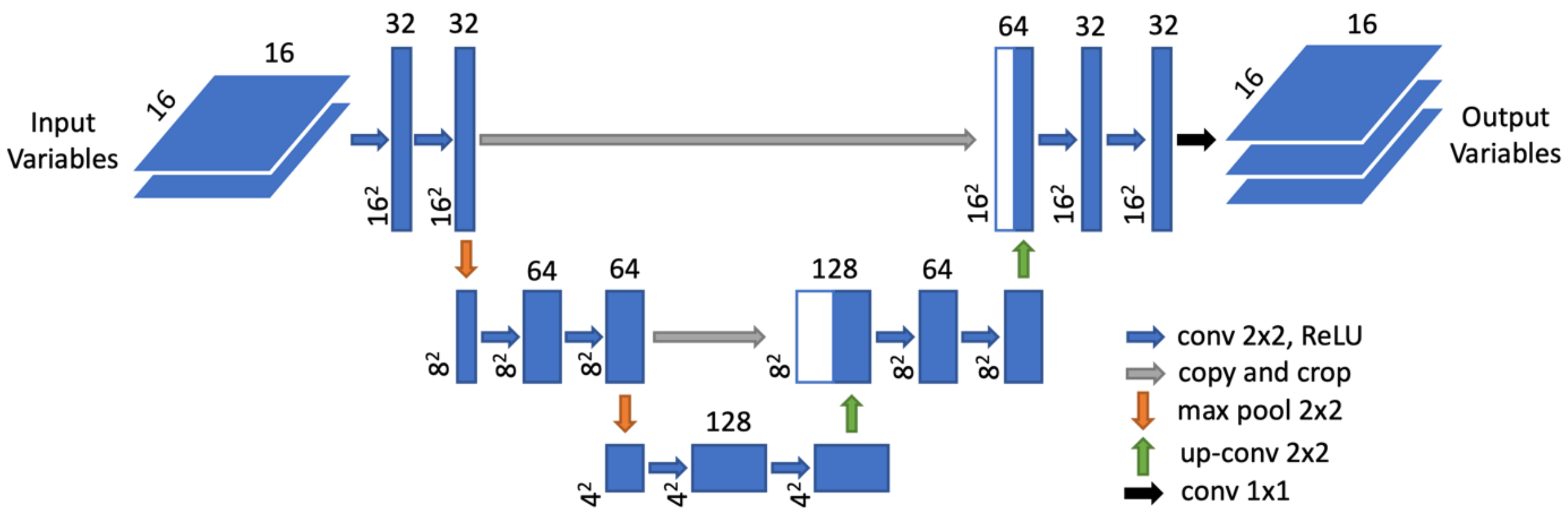

Figure 1 U-Net architecture with dimensions and notations following Ronneberger et al. (2015). This example corresponds to the training data from QG-L DA experiments, where input (2x16x16) and output (3x16x16) variables are localized potential vorticity and localized ensemble covariance matrices, respectively.

The contracting path of the U-Net (left side of Figure 1) consists of three applications of double 2x2 convolutions. Each convolution (blue arrow) is padded to maintain the same size between input and output images and includes a rectified linear unit (ReLU). The first two double convolutions are followed by 2x2 max pooling (orange arrow) operations with stride 2 for downsampling. Each step in the expansive path (right side of Figure 1) consists of a 2x2 up-convolution (green arrow) for upsampling, a concatenation with the corresponding high-resolution feature map from the contracting path (grey arrow), and double 2x2 convolution and ReLU (blue arrow). The final 1x1 convolution (black arrow) matches the number of channels in the output images. In this proof-of-concept paper, instead of a more comprehensive hyper-parameters sweep, we will focus on 2 main hyper-parameters in this U-Net architecture, the number of downsampling steps and the number of features after the first convolution, which are often referred to as the depth and width of the U-Net, respectively. The default U-Net as shown by Figure 1, has 2 downsampling steps and 32 features after the first convolution.

### 4.2. Data and methods

EnKF experiments of various lengths are performed in both QG-L and QG-H to generate the training data for the U-Net. These experiments are separate from those described in Section 2, but follow the same setup. A primary advantage of EnKF over 3DVar is the use of flow-dependent forecast error covariances that are estimated based on an ensemble of perturbed model simulations. Therefore, we would anticipate that the background model state (flow), even without the explicit ensemble perturbation statistics, could provide relevant information about the short-term forecast error covariances that vary over time.

The model ensemble-mean forecast potential vorticity (input for U-Net) and localized ensemble covariance matrices (output for U-Net) at each DA step are stored for training. Even for simple dynamic models such as the QG model used in this study, it is prohibitive to store the full covariance matrices for every model gridpoint at every DA step. Fortunately, due to the use of localization schemes in most ensemble DA systems, we only need to store localized covariance matrices that are large enough to cover the desired localization radius. Following the localized covariance matrices, the model ensemble-mean forecast potential vorticity is also localized around each model gridpoint to serve as the training data. One can also save all the ensemble perturbations and calculate the covariances during the training process. The relative efficiency between storing the localized covariance matrices or the ensemble perturbations will be case dependent.

In an EnKF experiment with QG-L (QG-H), there are 1024 (4096) paired training samples from each DA step, where each model location provides 1 paired sample of localized potential vorticity and localized ensemble covariance matrix. The data samples are divided into training samples and validation samples, which are 80% and 20% of the data, respectively. For training data from QG-L (Figure 1), each input sample has the shape 2x16x16, consisting of 2 model layers of localized potential vorticity around a central reference point, and each output sample has the shape 3x16x16, consisting of 3 localized covariance matrices that contain the covariances between each gridpoint and the same central reference point. The 3 matrices are the covariances for the 1$^{st}$ and 2$^{nd}$ model layers individually, as well as the cross covariances between the 1$^{st}$ and 2$^{nd}$ model layers.

The EnKF experiments that provide the training data are listed in Table 1. The 1280-member large-ensemble experiment DATA_L1280 with QG-L could minimize the sampling error to provide the best possible training data, while the more realistic DATA_L20 and DATA_L80 are used to test the robustness of the trained U-Net with more noisy data, and are also more representative of real-world ensemble sizes. In particular, DATA_L20 (and DATA_H20) is only 2 years to test the U-Net training with small ensemble size and limited data availability. The localization radius and relaxation factor of the training experiments do not require extensive tuning, as long as they maintain reasonable ensemble spread. We tested different DA parameters for the training experiments, which result in similar trained U-Nets. U-Nets are also trained with training data from EnKF experiments in QG-H to compare with transferred U-Nets in UNetKF experiments in QG-H. The application of the U-Nets listed in Table 1 in UNetKF experiments will be detailed in Section 5.

| | Numerical model for DA | Data length | Ensemble size | Localization radius | Relaxation factor | Trained U-Net |
|---|---|---|---|---|---|---|
| DATA_L20 | QG-L | 2 years | 20 | 100 km | 0.6 | UNet_L20 |
| DATA_L80 | QG_L | 9 years | 80 | 100 km | 0.5 | UNet_L80 |
| DATA_L1280 | QG-L | 9 years | 1280 | 100 km | 0.45 | UNet_L1280 |
| DATA_H20 | QG_H | 2 years | 20 | 100 km | 0.6 | UNet_H20 |
| DATA_H80 | QG-H | 9 years | 80 | 100 km | 0.45 | UNet_H80 |

Table 1 List of DA experiments that provide the training data.

### 4.3. U-Net Training and Evaluation

The training of the U-Net uses batch sizes ranging from 8000 to 32000 depending on the available graphical memory, a learning rate of 0.002, a root-mean-square error (RMSE) cost function, and the PyTorch Adam stochastic gradient descent optimizer. The choice of the training parameters is beyond the scope of this paper. Figure 2 shows the training and validation losses through 200 training epochs for UNet_L20, UNet_L80, and UNet_L1280. Although the training loss keeps decreasing generally as the training continues, the validation loss stagnates, or even increases after a certain number of epochs. Training is usually stopped when the validation loss stops decreasing to avoid overfitting to the training data. The minimal validation error occurs at epoch 81 for UNet_L20, epoch 74 for UNet_L80, and epoch 105 for UNet_L1280. We choose the U-Nets at the epochs of minimal validation errors to be used in later UNetKF experiments.

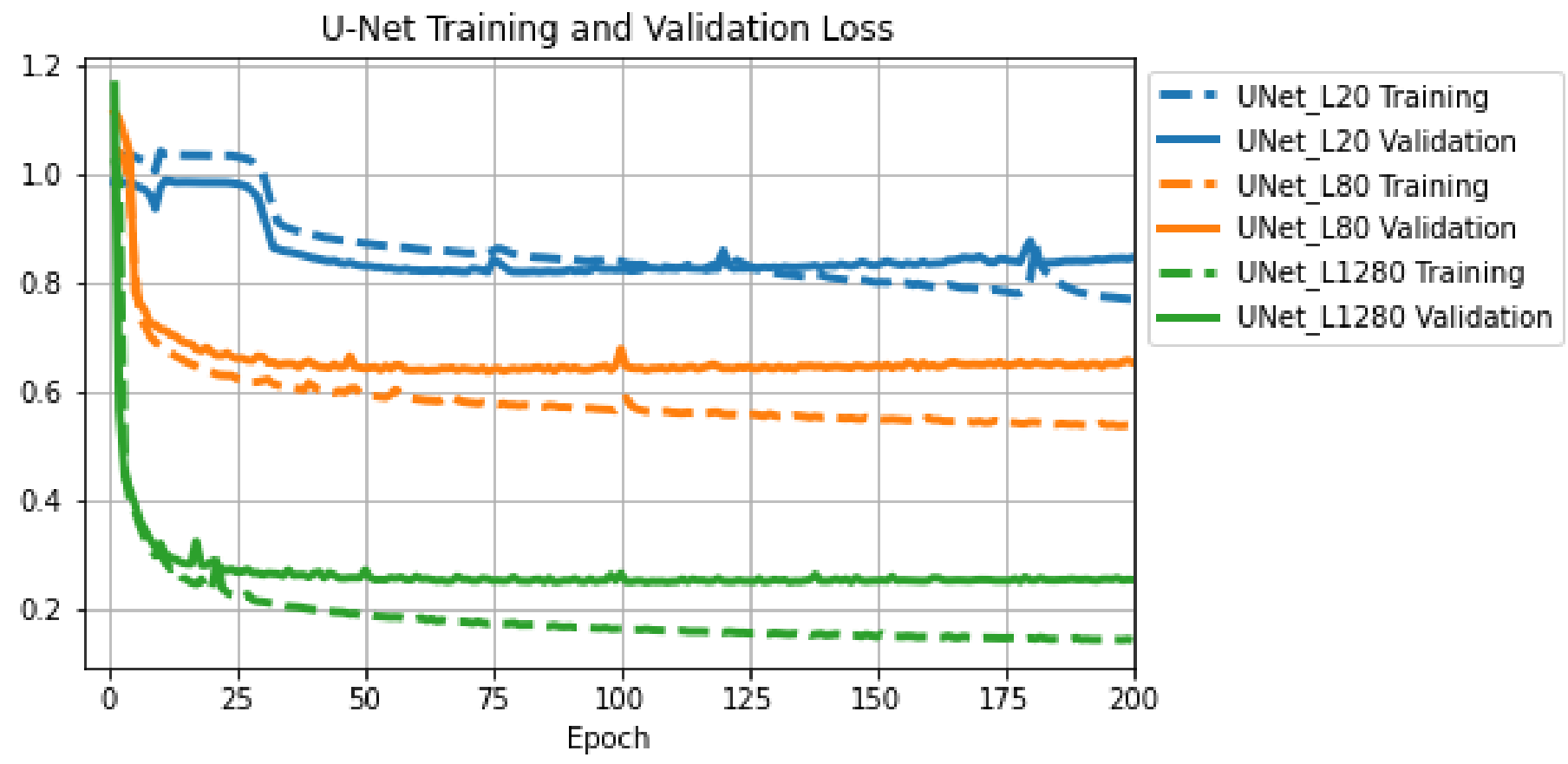

Figure 2 Training and validation losses during the training of UNet_L1 and UNet_L2.

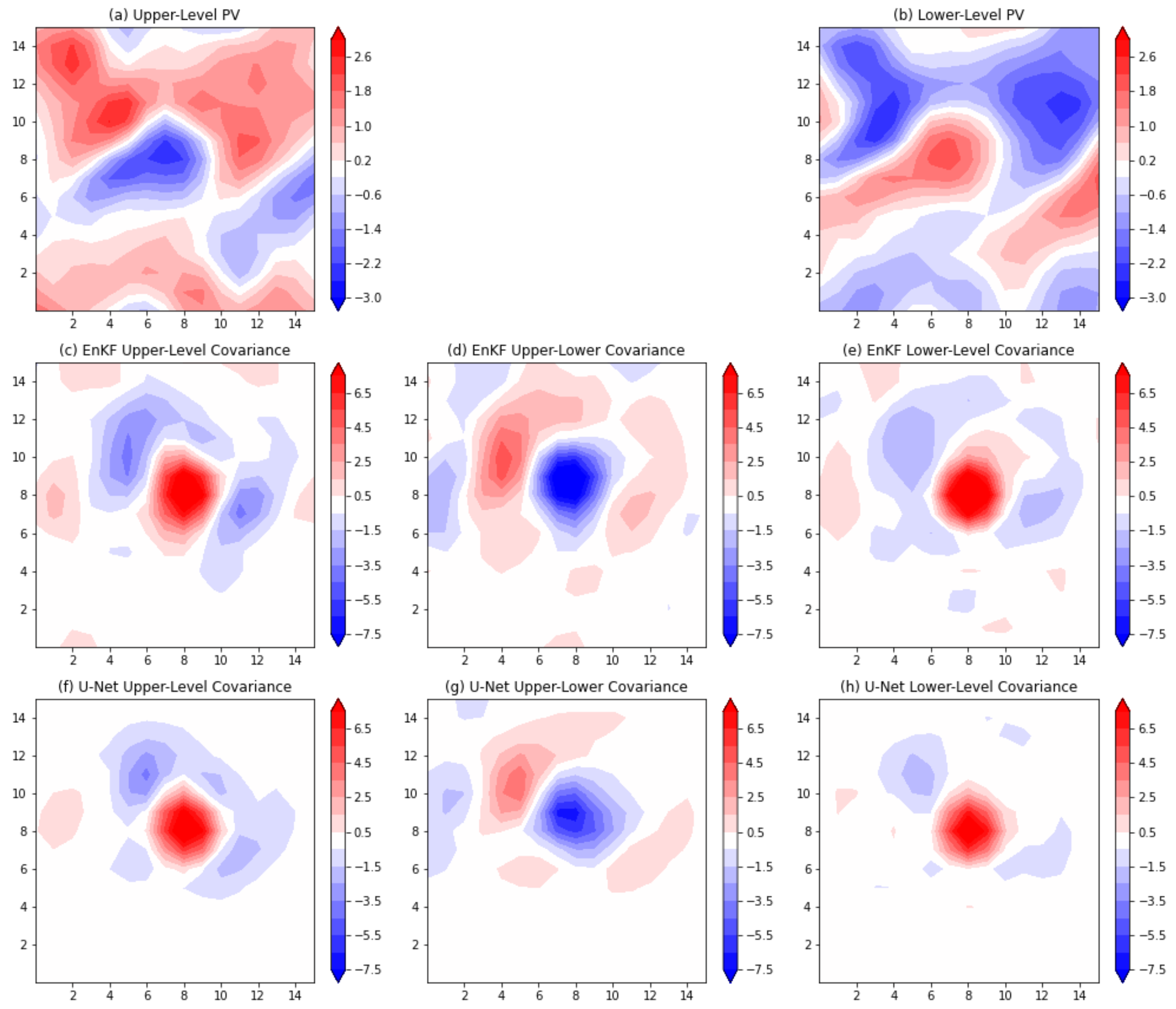

Figure 3 Example of UNet_L1280 for one validation data sample. (a, b) Upper- and lower-level potential vorticity, respectively. (c-e) EnKF ensemble covariances within the upper level, between the upper and lower levels, and within the lower level, respectively. (f-h) UNet_L1-predicted covariances given (a) and (b) as inputs, corresponding to (c-e), respectively.

It is notable that the training and validation errors decrease as the ensemble sizes increase in the training data from DATA_L20 to DATA_L1280. The primary reason is that the training data from DATA_L1280, particularly the ensemble covariances, have less “noise” than those from DATA_L20. The “noise” refers to the larger spurious sampling errors when calculating the ensemble covariances based on a smaller ensemble in DATA_L20 than in DATA_L1280. Figure 3 and Figure 4 show one example among the validation data samples for UNet_L1280 and UNet_L20, respectively. The sample includes the localized ensemble-mean potential vorticity for both levels (top panel), the localized EnKF ensemble covariances (middle panel) estimated based on the ensemble perturbations, and the U-Net predicted localized covariances (bottom panel) using the top panel as inputs. The potential vorticities in Figure 3a-b and Figure 4a-b look similar because they are ensemble-mean analysis around the same model gridpoint at the same time. Although the covariances around the central reference gridpoint show similar patterns (Figure 3c-e and Figure 4c-e), the ensemble covariances from DATA_L20 (Figure 4c-e) have larger non-zero values between the central and peripheral gridpoints, namely model locations that are physically far away from each other that we would not expect any correlated variability within one DA step, while DATA_L1280 (Figure 3c-e) could more accurately estimate zero or close-to-zero covariance values given its much larger ensemble size. The ensemble covariances close to the central gridpoint are also larger in DATA_L1280 than in DATA_L20. Therefore, during training, DATA_L1280 provide higher-fidelity covariance output data for UNet_L1280 to learn from, while the input data are similar between UNet_L1280 and UNet_20. Subsequently, during validation, the “noise” in the validation samples of DATA_L20 cannot and should not be predicted by the U-Net, which leads to larger validation errors for UNet_L20. However, because these spurious covariances are caused by sampling errors and change randomly over time, UNet_L20 is able to filter out these random noise and makes predictions (Figure 4f-h) that resemble those from UNet_L1280 (Figure 3f-h) despite the larger noise in its training data, albeit with smaller values that are in line with the smaller values in DATA_L20 compared to DATA_L1280. Although the localized covariance matrices in the training data are not subject to the Gaspari-Cohn localization weights, the trained U-Nets themselves are able to function as localization schemes by filtering out spurious covariances in the training data and predict flow-dependent error covariances close to each model gridpoint. This U-Net equivalent of localization is also “adaptive” by nature since the predicted covariances are conditioned on the background model state.

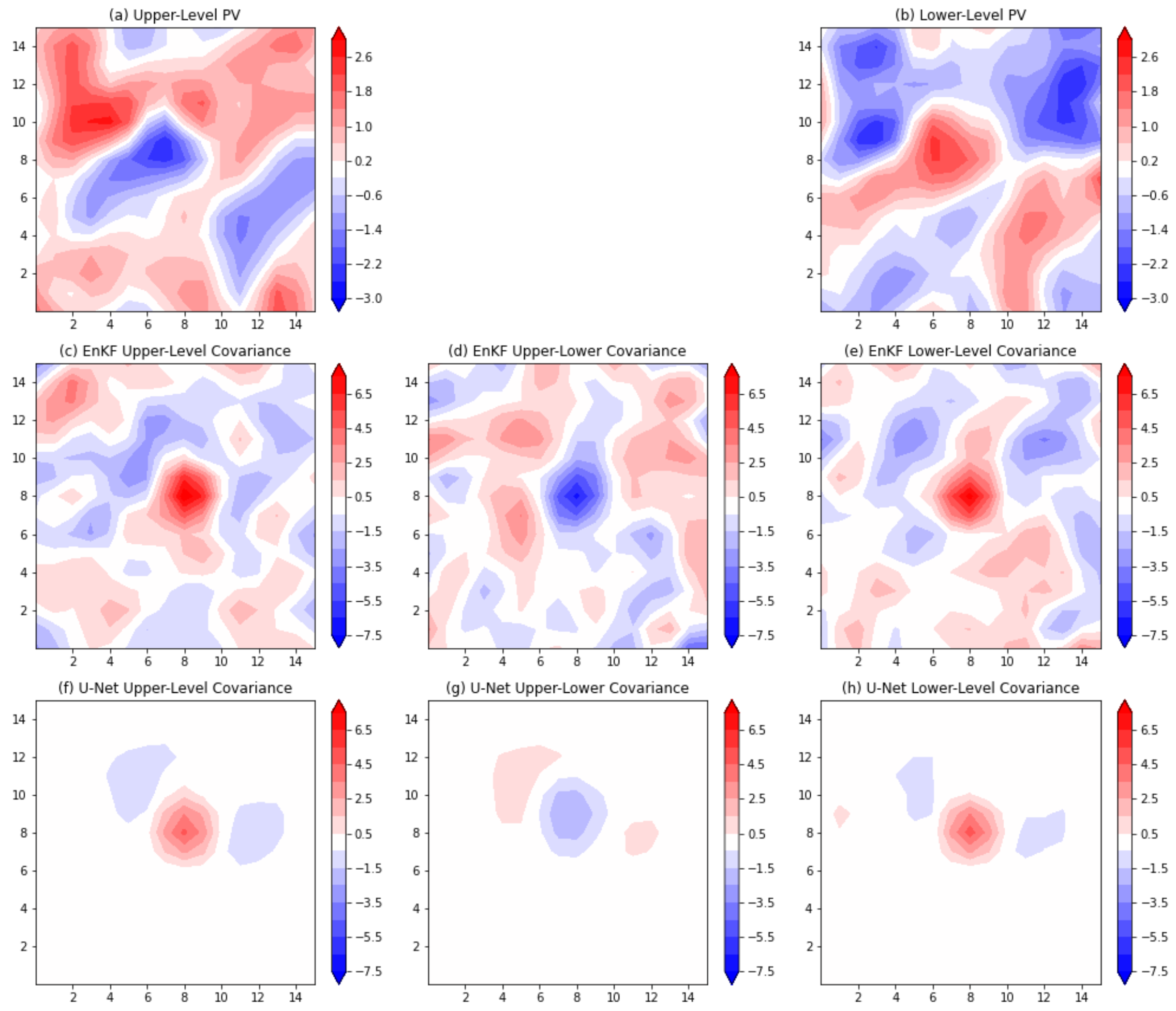

Figure 4 Same as Figure 3, but for UNet_L20.

To evaluate the skills of the U-Nets across all model gridpoints and data assimilation steps, we applied UNet_L1280, UNet_L80, and UNet_L20 to the full DATA_L1280 dataset. Here, we consider the 1280-member-estimated covariances as the “true covariances”, and Figure 5 shows the average changes in RMSE against the *time-varying* “true covariances” across all model gridpoints when the U-Net-predicted covariances are used instead of the *time-averaged* *climatological* “true covariance”. In other words, any values below 1.0 in Figure 5 means that the U-Net predictions provide more useful covariances than the climatological “true covariance”. As expected, UNet_L1280 (Figure 5a-c) provides the most reduction in RMSE, followed by UNet_L80 (Figure 5d-f). UNet_L20 (Figure 5g-i) is able to beat the climatological “true covariance” within certain localization distances for the first layer, and between the first and second layers, but not for the second layer. This could be attributed to the increasing sampling errors in the DATA_L20 training data due to its limited ensemble size and shorter experiment length. Although the reduction in RMSE is small for UNet_L20, it should be noted that the “true covariances” from the 1280-member DA experiment are themselves good estimations of the model’s uncertainty, and are not available for typical DA systems. The static error covariances of 3DVar lead to increased RMSE compared to the climatological “true covariances” for most of the localized domain, because the static covariances are calculated in a different method that leads to different covariance magnitude and patterns.

The impact of the depth and width hyper-parameters of the U-Net is also tested. Large depth or width might potentially improve the skill of the U-Net, but at the expense of more computational costs for both training and inference, as well as higher risk of overfitting. The overfitting issue arises when the neural network is excessively trained to fit the high-frequency or small-scale “noisy” information from the training data, which may not benefit its prediction skill outside of the training samples. This issue usually manifests itself during training in the form of stagnant or increasing validation loss even when the training loss keeps decreasing. Larger neural networks could be more prone to overfitting since there are more parameters to be trained to fit the “noise” in the training samples. We tested all combinations between widths of 16, 32 or 64 and depths of 2 or 3 (not shown). Although the training loss generally decreases when the U-Net is wider or deeper, the minimal validation loss does not always follow, and the gap between training and validations losses usually increases, indicating that the wider or deeper U-Net may have been overfitted to the training samples and will have lower predictive skills outside of the training samples. The default U-Net with a depth of 2 and width of 32 yields the smallest validation error.

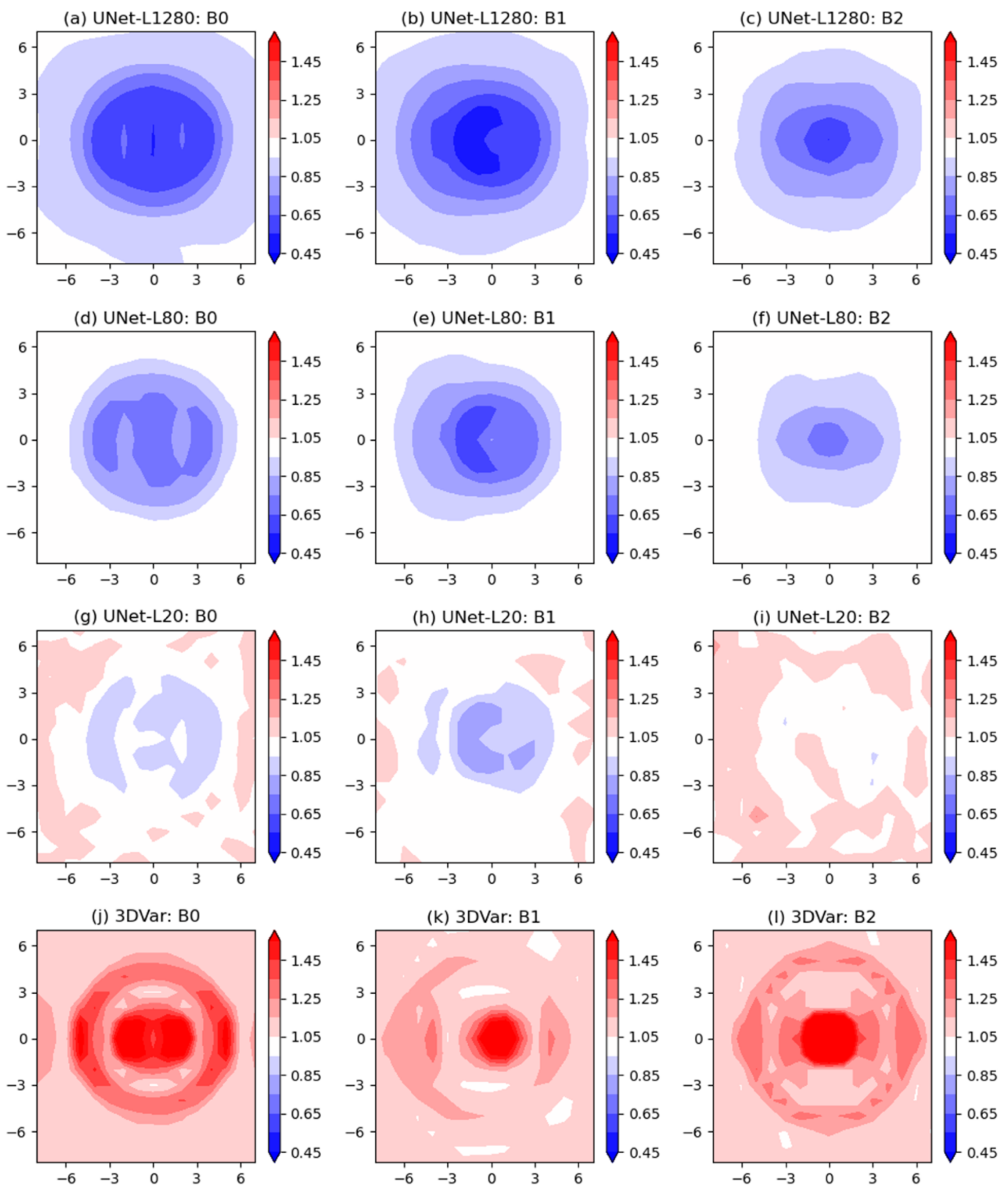

Figure 5 The accuracy of localized error covariance estimations from (a-c) UNet_L20, (d-f) UNet_L80, (g-i) UNet_L1280 and (j-l) 3DVar across all model gridpoints in QG-L. The values in the plots indicate the ratio of RMSE, calculated from the errors against the time-varying error covariances of DATA_L1280, between U-Net/3DVar estimations and DATA_L1280's own climatological covariance matrices.

## 5. UNetKF results

The performance of UNetKF is first compared to 3DVar, En3DVar, and EnKF in a series of DA experiments using QG-L as the forecast model. The results of the experiments are shown in Figure 6, while more details about the experiments are listed in Table S1. The localization radius and relaxation parameter are selected to minimize the RMSE of each experiment. The U-Nets used here are UNet_L20, UNet_L80 and UNet_L1280, which are trained with data from the same QG-L model. The colors of the bars indicate the DA method (3DVar, En3DVar, EnKF, UNetKF) or the specific U-Net that is used by UNetKF. The ensemble sizes in Figure 6 refer to the DA experiments here and are not related to the ensemble sizes of the U-Net training data. Each U-Net, regardless of the ensemble size of its training data (e.g. 20 for UNet_L20, 80 for UNet_L80), is applied in UNetKF experiments of various ensemble sizes ranging from 1 to 80. The ensemble-mean RMSE and ensemble standard deviation (spread) for both model layers are shown in Figure 6 for all QG-L experiments when available. Since the relative performance among the different DA experiments holds for both layers in most cases, we will not distinguish the two layers when discussing the results in this section unless specified otherwise.

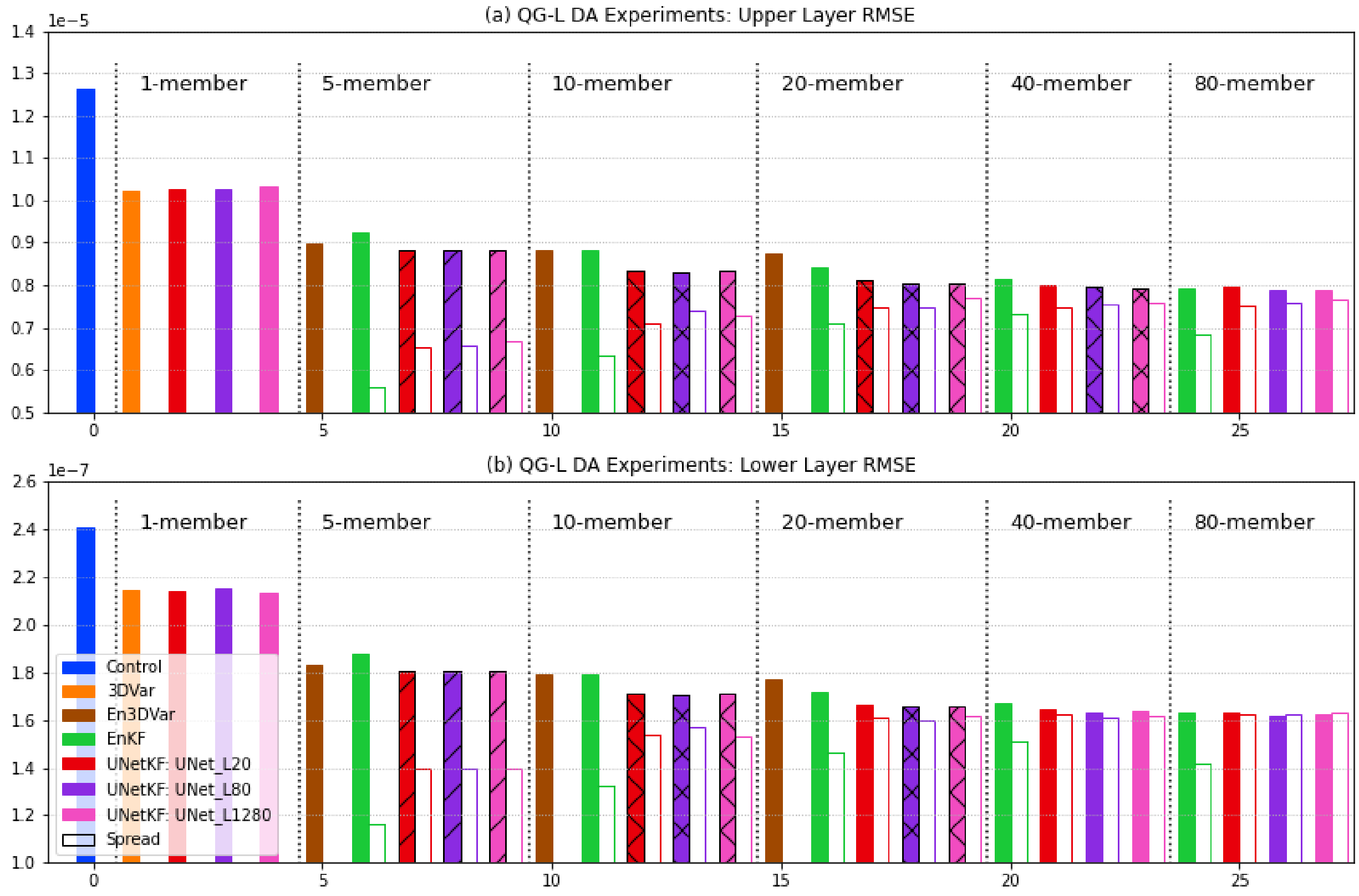

Figure 6 Analysis RMSE and ensemble spread (when available) of (a) upper-level and (b) lower-level potential vorticity for control and DA experiments in the QG-L model. The filled and unfilled bars indicate the RMSE and spread, respectively. The ensemble sizes are listed in the plot for each group of 3DVar, En3DVar, EnKF and UNetKF experiments. The cross-hatched bars indicate that RMSE of UNetKF is significantly smaller than those of both EnKF and En3DVar for the same ensemble size, while the diagonal-hatched bars indicate that RMSE of UNetKF is significantly smaller than that of EnKF.

When implemented in UNetKF experiments, UNet_L20, UNet_L80 and UNet_L1280 result in similar performance across different ensemble sizes despite their gap in training and validation errors as discussed earlier. UNetKF with UNet_L20 performs slightly worse than UNetKF with UNet_L80 or UNet_L1280, but the differences are not significant for any specific ensemble size. This means that UNet_L20 can successfully filter out the “noise” in the training data from DATA_L20. This is promising for future applications of UNetKF, since we can train the U-Net with covariance data based on moderate ensemble sizes that are practical for more computationally expensive dynamic models. Between En3DVar and EnKF, En3DVar outperforms EnKF with 5 members, but EnKF catches up with 10 members and outperforms En3DVar with more than 10 members.

With a single member, UNetKF experiments have similar RMSE to 3DVar, where the differences are not significant between 3DVar and any UNetKF experiment. With 5 members, UNetKF experiments significantly outperform EnKF, while their outperformance to En3DVar is not significant. With 10 or 20 members, all UNetKF experiments significantly outperforms both En3DVar and EnKF, except for the 2$^{nd}$ layer of 20-member UNetKF experiment with UNet_L20. EnKF gradually catches up to UNetKF with 40 or 80 members, where UNetKF still outperforms EnKF, but the differences are not significant, except for the 1$^{st}$ layer of 40-member UNetKF experiments with UNet_L80 and UNet_L1280.

In sum, the QG examples in Figure 6 show that UNetKF can match or exceed the performance of 3DVar, En3DVar or EnKF for typical ensemble sizes used in weather and climate applications. In particular, the significant advantage of UNetKF with smaller ensemble sizes (5-20) compared to EnKF or En3DVar is promising for potential applications in more complex models, of which we may only afford to run small ensembles. The outperformance of UNetKF is more pronounced for smaller ensemble sizes because it does not rely on the model ensemble statistics to accurately estimate the covariance matrices like in EnKF. As shown in Section 4, the U-Net is effective in filtering out the “noise”, or spurious covariances from the training data, which were not subject to the Gaspari-Cohn localization weights from the EnKF experiments. As a result, the U-Nets could provide their own equivalent of “adaptive localization” to UNetKF and make UNetKF less sensitive to the choices of the ensemble inflation (relaxation) and localization parameters than EnKF. Although the inflation and localization parameters were fixed for the training experiment (e.g. DATA_L20 or DATA_L80), the training of U-Nets do not require the optimal EnKF parameters, as long as the experiments maintain healthy ensemble spread to sample the forecast covariances.

Figure 7 shows the RMSE sensitivities on the relaxation factors and localization distances for the 20-member EnKF experiments and the 20-member UNetKF experiments with UNet_L80, respectively. In the case of EnKF (Figure 7a, b), there is a narrow range of parameter choices that yield the minimal RMSEs, and the optimal choices also differ for the two model levels due to their differences in the spatial and temporal scales of variability. UNetKF (Figure 7c, d) not only significantly outperforms EnKF in terms of the minimal RMSEs, but also achieve the minimal or close-to-minimal RMSEs across wider ranges of relaxation factors and localization radii. These behaviors of parameter sensitivities persist across different ensemble sizes for EnKF and UNetKF in both the QG-L and QG-H models (not shown). Although the training experiments still require some tuning of the DA parameters, they can be achieved with shorter sensitivity experiments and do not require the optimal parameters for minimal RMSE. This feature of UNetKF would accelerate tuning of DA parameters in complex models, for which parameter sensitivity sweeps like Figure 7 are prohibitively expensive in some cases.

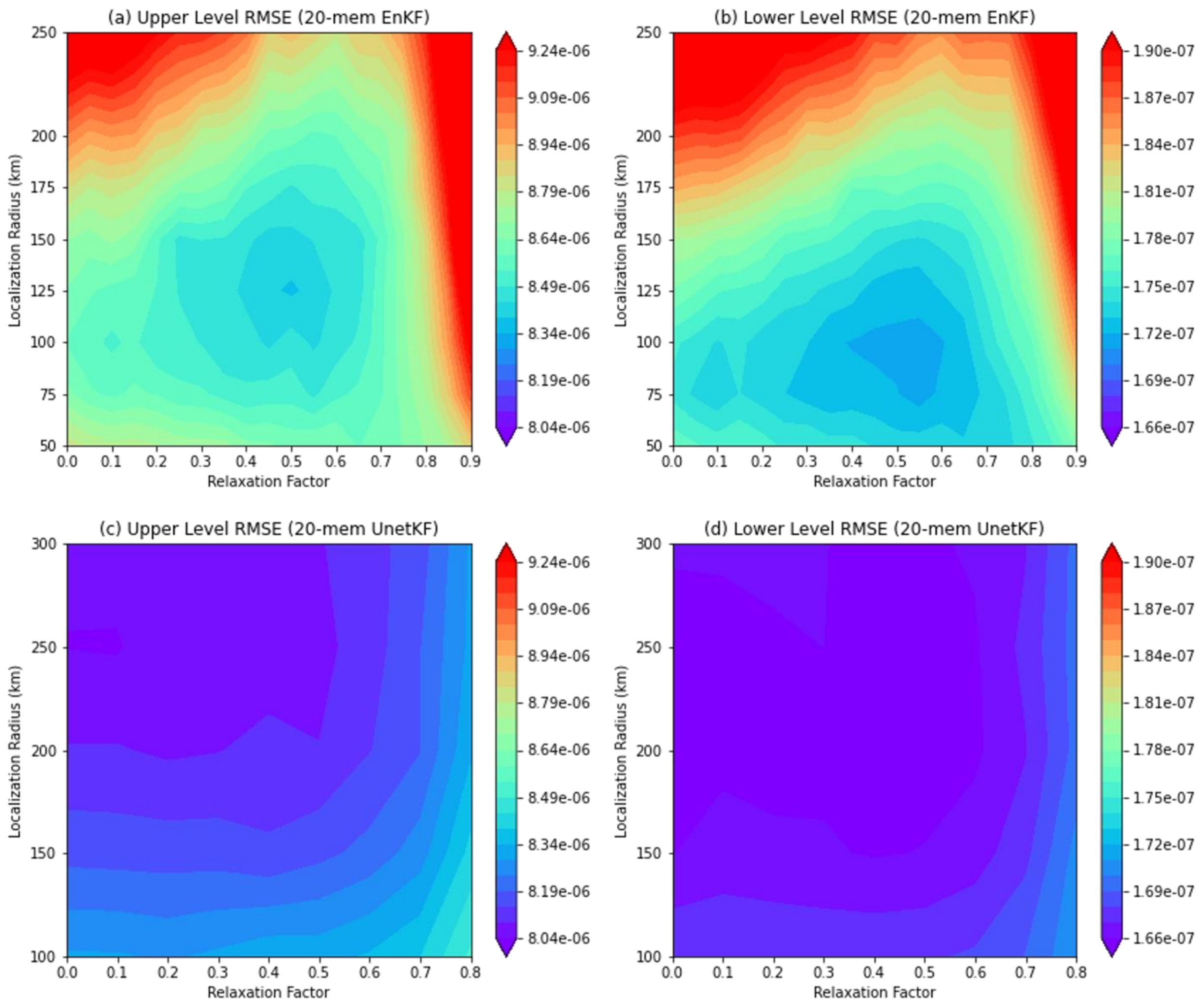

Figure 7 Dependence of RMSE on the choices of localization radius and relaxation factor for (a) upper level and (b) lower level of 20-member EnKF experiments, and (c) upper level and (d) lower level of 20-member UNetKF experiments with UNet_L80.

One argument against the UNetKF experiments shown in Figure 6 is that the training of the U-Net requires a working ensemble-based DA system in the same DA model to provide the training data. This may not be feasible in real world applications where building a full DA system and saving the necessary covariance matrices are impractical. In other words, if there already exists an ensemble-based DA system for the target DA model, despite the outperformance of UNetKF with smaller ensemble sizes compared to EnKF, one might as well just use the existing DA system instead of going through the process of saving all the covariance matrices and training U-Nets.

Given that the U-Nets are trained to learn and predict ensemble covariances that reflect the variability of the simulated dynamic system, one might expect them to be transferrable between models that simulate the same underlying dynamic system, such as the same GCM with different resolutions, or different GCMs for the same earth system component.

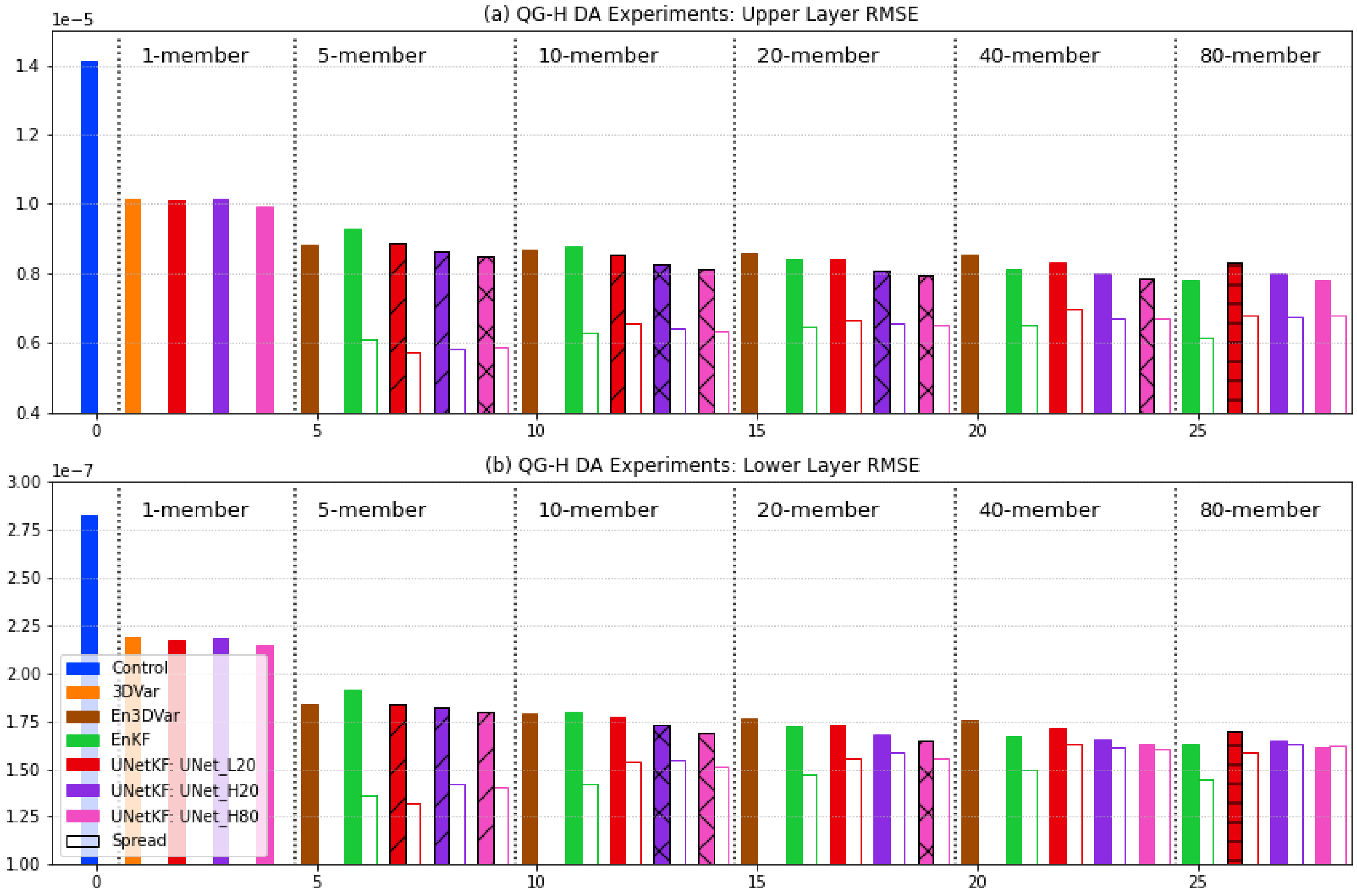

Figure 8 Same as Figure 6, but for experiments in the QG-H model. The horizontal-hatched bars indicate that UNetKF is significantly worse than EnKF for the same ensemle size.

Therefore, DA experiments using QG-H as the DA forecast model are performed to test the feasibility of UNetKF using trasnferred U-Nets. Figure 8 shows DA experiments in QG-H using 3DVar, En3DVar, EnKF, and three sets of UNetKF with UNet_L20, UNet_H20 and UNet_H80, respectively (Table S2). UNet_H20 and UNet_H80 are trained with the data from DATA_H20 and DATA_H80, respectively, which are performed with the same QG-H model, while UNet_L20 is the same U-Net used in Figure 6. Since UNet_L20 is trained with lower-resolution data from QG-L, two additional steps are required to apply it to UNetKF experiments in QG-H. First, the QG-H model-forecasted potential vorticity fields at DA steps are subsampled to the lower QG-L resolution to serve as inputs for UNet_L20. Then, the UNet_L20-predicted covariance matrices are upsampled back to the higher QG-H resolution to be used in the Kalman gain formula. For the subsampling, methods including bilinear interpolation, cubic interpolation and area-weighted averaging produce similar results. Different upsampling methods including bilinear and cubic interpolations also produce similar results. UNet_H20 and UNet_H80 are used here as the benchmark to gauge the performance penalty due to the use of a transferred UNet_L20 from QG-L to QG-H.

With UNet_H20 and UNet_H80 that are trained with native QG-H data, the relative performance among UNetKF, 3DVar, En3DVar and EnKF in QG-H (Figure 8) are similar to their QG-L counterparts (Figure 6). More specifically, En3DVar is able to outperform EnKF with 5 or 10 members, but gets passed by EnKF for 20 or more members. UNetKF with UNet_H20 or UNet_H80 significantly outperforms En3DVar with 10 or 20 members, and significantly outperforms EnKF with 5, 10 or 20 members for all RMSE except the lower layer in one experiment. UNetKF with a single member matches the 3DVar performance without significant differences in RMSE.

Comparing among the UNetKF experiments, the ones using UNet_L20 have expectedly higher RMSE than the ones with UNet_H20, but the RMSE is only 3-5% larger than the ones using UNet_H20 across different ensemble sizes. Even with the performance penalty of the transferred UNet_L20, UNetKF is still able to significantly outperform EnKF with 5 or 10 members. This bodes well for future applications, given that we might only afford to run ensembles with sizes in this range for future high-resolution ocean or climate reanalysis systems. As the ensemble size becomes larger, the performance of EnKF improves faster, eventually becomes significantly better than UNetKF with the transferred UNet_L20 at 80 ensemble members. However, UNetKF with the natively-trained UNet_H20 and UNet_H80 is still comparable to EnKF at 80 members.

The relative small penalty of using UNet_L20 compared to UNet_H20, combined with the comparison with EnKF, confirms that the transferred UNet_L20, despite being trained with data from a lower-resolution model, is able to partially capture the flow-dependent error covariances of the higher-resolution model, given that both models are intended to simulate the same underlying dynamic system.

## Summary and discussion

In this paper, we use U-Net, a type of convolutional neural network, to predict the localized error covariances for ensemble-based data assimilation methods. Using the 2-layer QG models, U-Nets are trained using data from EnKF experiments, where the input and output are the localized ensemble-mean potential vorticity and the localized ensemble covariance matrices, respectively. In UNetKF experiments, the flow-dependent covariance matrices are predicted by U-Nets instead of being estimated from model ensembles in EnKF. First implemented in a lower-resolution QG-L model, UNetKF performs as well as or better than 3DVar, En3DVar and EnKF with ensemble sizes up to 80, while being less sensitive to choices of DA parameters for ensemble inflation and covariance localization. Furthermore, we successfully transferred trained U-Nets to a higher-resolution QG-H model for UNetKF experiments, which again performs competitively to EnKF. The performance of UNetKF with small to moderate ensemble sizes could be advantageous for computationally expensive dynamic models. The possibility of transferring trained U-Nets across models also helps facilitate the implementation of UNetKF in cases when the target DA forecast model does not have an existing ensemble-based DA system, or when it is not practical to train the U-Net with native-resolution data.

Computational efficiency is a huge advantage of ML methods in the cases of data-driven weather forecast or emulation models. However, the exact time saving of UNetKF compared to EnKF would depend on the computational costs of the dynamic models in which UNetKF is implemented, since the efficiency advantage would come from the possibility of running less ensemble members without sacrificing performance. Although this advantage is not highly prominent here given the low computational cost of the QG model, for more computationally expensive models, the saving would be outstanding if a few less ensemble members are needed. There are also some complications to the efficiency question. First, the efficiency of U-Net inference depends on the availability of graphic processing units (GPUs). Without GPUs, UNetKF would take longer than EnKF for the same ensemble size on the same number of central processing unit (CPU) cores. With the additional GPUs, the U-Net inference time can be reduced several folds and make UNetKF faster than EnKF. However, there is also the possibility that traditional EnKF methods could also be performed on GPUs to improve efficiency.

Although UNetKF shows promising results as an example of ML-assisted ensemble DA, there is certainly room for improvement in some aspects of this study for future research. First, the performance of UNetKF with a single member is only comparable to 3DVar. Further improvement may be possible with changes in the DA formulation. For example, ML methods might enable the use of 4DVar-like methods without the need for tangent linear and adjoint models (Fablet et al., 2021; Lafon et al., 2023). Second, the improvement in how to transfer the U-Net to different models, beyond just statistical interpolation in this study, might be possible with additional considerations for the models' dynamics and statistics, or additional ML methods. Last but not the least, the scope of applications could be greatly broadened if the U-Net can be trained without the need for existing ensemble DA systems to provide the training data. It will be interesting to explore whether the ensemble error covariances can be retrieved from (ensemble) control simulations.

Despite the successful implementation of UNetKF in the QG model in this study, some important questions need to be answered before we can apply UNetKF, or other ML-assisted DA to more complex models, especially state-of-the-art climate models. First, the amount of data and computation required to train and implement ML models in GCMs might be a substantial obstacle. The solution would require more collaboration with data and computer scientists to build the appropriate hardware and software infrastructure, however, this aligns well with the goals of incorporating more ML/AI methods in climate models for the community. Second, like other efforts that try to apply ML/AI methods in climate models, there are questions such as how to handle coasts and topography, the out-of-sample issue for climate change, etc. Last but not the least, given the rapid progress in ML methodology, better architectures may already exist or will emerge that could improve upon the current study in terms of accuracy and/or efficiency. Nonetheless, this proof-of-concept study demonstrates exciting possibilities for ML/AI to advance data assimilation research and applications.

**Acknowledgement**

This study is supported by the Stellar computing cluster at Princeton University. I would like to thank my collaborators in the M2LInES (Multiscale Machine Learning In Coupled Earth System Modeling) project for inspirations and discussions about various machine learning topics.

**Open Research**

PyQG is publicly available at https://github.com/pyqg/pyqg. All code for this study is available at https://github.com/feiyulu/pyqg_DA.

Supporting Information for

U-Net Kalman Filter (UNetKF): An Example of Machine Learning-assisted Ensemble Data Assimilation

Feiyu Lu[1, 2, •]

[1]University Cooperation of Atmospheric Research

[2]National Oceanic ana Atmospheric Administration/Geophysical Fluid Dynamics Laboratory

[*]Previously Princeton University

Contents of this file

| DA Method | Ensemble Size | U-Net | Relaxation Factor | Localization Radius (km) | RMSE | |
|---|---|---|---|---|---|---|
| | | | | | Upper lvl ($10^{-6}$) | Lower lvl ($10^{-7}$) |
| Control | 1 | N/A | N/A | N/A | 12.6 | 2.41 |
| 3DVar | 1 | | | 100 | 10.2 | 2.15 |
| En3DVar | 5 | | | | 8.97 | 1.83 |
| | 10 | | | | 8.82 | 1.80 |
| | 20 | | | | 8.74 | 1.78 |
| EnKF | 5 | | 0.6 | 50 | 9.26 | 1.88 |
| | 10 | | 0.55 | 75 | 8.81 | 1.79 |
| | 20 | | 0.55 | 100 | 8.43 | 1.72 |
| | 40 | | 0.5 | 125 | 8.17 | 1.67 |
| | 80 | | 0.35 | 175 | 7.94 | 1.63 |
| UNetKF | 1 | UNet_L20 | N/A | 50 | 10.3 | 2.14 |
| | 5 | | 0.1 | 150 | 8.82 | 1.81 |
| | 10 | | 0.0 | 200 | 8.34 | 1.71 |
| | 20 | | 0.2 | 250 | 8.13 | 1.67 |
| | 40 | | 0.0 | 250 | 8.00 | 1.64 |
| | 80 | | 0.0 | 250 | 7.96 | 1.63 |
| | 1 | UNet_L80 | N/A | 50 | 10.3 | 2.15 |
| | 5 | | 0.1 | 150 | 8.83 | 1.81 |
| | 10 | | 0.3 | 250 | 8.32 | 1.71 |
| | 20 | | 0.0 | 250 | 8.04 | 1.66 |
| | 40 | | 0.1 | 250 | 7.95 | 1.63 |
| | 80 | | 0.1 | 250 | 7.89 | 1.62 |
| | 1 | UNet_L1280 | N/A | 50 | 10.3 | 2.13 |
| | 5 | | 0.2 | 150 | 8.81 | 1.81 |
| | 10 | | 0.2 | 200 | 8.33 | 1.71 |
| | 20 | | 0.3 | 250 | 8.04 | 1.66 |
| | 40 | | 0.1 | 250 | 7.94 | 1.64 |
| | 80 | | 0.1 | 250 | 7.88 | 1.63 |

Table S1 The list of DA experiments in QG_L and their configurations.

| DA Method | Ensemble Size | U-Net | Relaxation Factor | Localization Radius (km) | RMSE | |
|---|---|---|---|---|---|---|
| | | | | | Upper lvl ($10^{-6}$) | Lower lvl ($10^{-7}$) |
| Control | 1 | N/A | N/A | N/A | 14.1 | 2.83 |
| 3DVar | 1 | | | 100 | 10.2 | 2.19 |
| En3DVar | 5 | | | | 8.82 | 1.84 |
| | 10 | | | | 8.68 | 1.79 |
| | 20 | | | | 8.58 | 1.77 |
| | 40 | | | | 8.53 | 1.75 |
| EnKF | 5 | | 0.7 | 50 | 9.29 | 1.91 |
| | 10 | | 0.6 | 75 | 8.78 | 1.80 |
| | 20 | | 0.55 | 100 | 8.42 | 1.73 |
| | 40 | | 0.5 | 125 | 8.11 | 1.67 |
| | 80 | | 0.45 | 200 | 7.82 | 1.63 |
| UNetKF | 1 | UNet_L20 | N/A | 50 | 10.1 | 2.17 |
| | 5 | | 0.1 | 100 | 8.87 | 1.84 |
| | 10 | | 0.1 | 200 | 8.55 | 1.77 |
| | 20 | | 0.0 | 150 | 8.40 | 1.73 |
| | 40 | | 0.2 | 200 | 8.33 | 1.71 |
| | 80 | | 0.0 | 150 | 8.29 | 1.70 |
| | 1 | UNet_H20 | N/A | 50 | 10.1 | 2.18 |
| | 5 | | 0.1 | 200 | 8.63 | 1.82 |
| | 10 | | 0.0 | 200 | 8.25 | 1.73 |
| | 20 | | 0.1 | 200 | 8.10 | 1.68 |
| | 40 | | 0.1 | 200 | 8.00 | 1.66 |
| | 80 | | 0.0 | 200 | 7.97 | 1.65 |
| | 1 | UNet_H80 | N/A | 100 | 9.94 | 2.15 |
| | 5 | | 0.1 | 200 | 8.51 | 1.79 |
| | 10 | | 0.0 | 200 | 8.13 | 1.69 |
| | 20 | | 0.1 | 200 | 7.95 | 1.65 |
| | 40 | | 0.2 | 250 | 7.85 | 1.63 |
| | 80 | | 0.0 | 250 | 7.80 | 1.62 |

Table S2 Same as Table S1 but for DA experiments in QG-H.